\documentclass[review]{elsarticle}
\usepackage{lineno,hyperref}
\modulolinenumbers[5]
\usepackage{array}
\usepackage{tikz}
\usepackage{pgfplots}

\usepackage{array}
\usepackage{tikz}
\usepackage{pgfplots}
\usepackage{algorithm}
\usepackage{algpseudocode}
\usepackage{amssymb}
\usepackage{latexsym}
\journal{Pattern Recognition Letters}
\pgfplotsset{width=10cm,compat=1.9}

\begin{document}

\begin{frontmatter}

\title{Robust Stereo Feature Descriptor for Visual Odometry}

\author{Ehsan Shojaedini}

\author{Reza Safabakhsh}
\address{shojaedini@aut.ac.ir, safa@aut.ac.ir }
\address{Amirkabir university\\
	Tehran, Iran\\}

\begin{abstract}
	In this paper, we propose a simple way to utilize stereo camera data to improve feature descriptors. Computer vision algorithms that use a stereo camera require some calculations of 3D information. We leverage this pre-calculated information to improve feature descriptor algorithms. We use the 3D feature information to estimate the scale of each feature. This way, each feature descriptor will be more robust to scale change without significant computations. In addition, we use stereo images to construct the descriptor vector. The Scale-Invariant Feature Transform (\textit{SIFT}) and Fast Retina Keypoint (\textit{FREAK}) descriptors are used to evaluate the proposed method. The scale normalization technique in feature tracking test improves the standard \textit{SIFT} by 8.75\% and improves the standard \textit{FREAK} by 28.65\%. Using the proposed stereo feature descriptor, a visual odometry algorithm is designed and tested on the \textit{KITTI} dataset. The stereo \textit{FREAK} descriptor raises the number of inlier matches by 19\% and consequently improves the accuracy of visual odometry by 23\%. \\\\
	\textbf{Keywords:} Stereo Camera, Feature Descriptor, Visual Odometry, Feature Matching, Depth Information, Stereo Visual Odometry.
\end{abstract}

\end{frontmatter}

\section{Introduction}

Feature descriptors presently constitute one of the fundamental components in most of the computer vision tasks. They have a key role in many vision algorithms such as scene reconstruction, visual odometry, object detection, and object recognition. As a result, any improvement in feature descriptors will probably increase the performance of such algorithms.

In visual odometry, feature descriptors are used to match features between displacements in the camera position. Feature matching is a very delicate work where each feature is a projection of a geometric point of the 3D scene to the image plane so that it would be recognizable from different views. Here, feature descriptors are used to match features between the left and right camera images and to track the features over time. There may be some incorrect matches among the matched features called outliers. These outliers are detected and rejected in an outlier rejection step of the visual odometry algorithm, and only the inlier matches are used. Moreover, not all of the inlier matches are suitable for motion estimation. In other words, some features are matched with a small drift and the feature descriptors are not sensitive enough to detect that. Although one-pixel error causes a small drift in the estimation, such drifts are accumulated and cause a notable error in the estimated path. Considering these issues, finding the best matches is considered as one of the main challenges in visual odometry.

In the recent years, many successful feature descriptors have been proposed such as: \textit{SIFT} by \cite{Lowe1999}, Speeded Up Robust Features (\textit{SURF}) by \cite{Bay2006}, Binary Robust Independent Elementary Features (\textit{BRIEF}) by \cite{Calonder2010}, Oriented FAST and Rotated BRIEF (\textit{ORB}) by \cite{Rublee2011}, and \textit{FREAK} by \cite{Alahi2012}. Some of them like \textit{SIFT} or \textit{SURF} are robust, but slow. Others like \textit{BRIEF}, \textit{ORB}, and \textit{FREAK} are fast, but sensitive to large transforms. Although they all have been successful in many application, they generally have a large mismatch rate. For example, in visual odometry, matched features are usually contaminated with outliers by more than 40\%. This happens when the feature descriptor is not discriminative at all points, especially the challenging ones. Therefore, an outlier rejection scheme (such as Random Sample Consensus (\textit{RANSAC}) by \cite{Fischler1981}) should be used to determine which one is a correct and which one is an incorrect match. A better solution will result if these mismatches do not occur. A solution is to prevent these mismatches from happening using more robust feature descriptors. 

In this paper, we propose a robust stereo feature descriptor with two properties: first, it utilizes depth information extracted from the stereo images to estimate the scale. Second, it uses the two images of the stereo camera to construct the feature descriptor vector. Our purpose is to build a feature descriptor that is both robust and efficient. A feature descriptor is calculated from a selected area around the corresponding feature. We leverage the depth information to change the area adaptively. Using the left and right images of the scene points helps us to have more information about the features and to implicitly check the features matching process. With the help of the depth information of the feature, we can correctly determine the distance between the feature and the camera. The smaller the distance from the camera is, the larger the area around the feature needed to construct the descriptor is. In other words, our stereo descriptor uses the depth information to estimate the correct scale for the features and adjust the scale for the descriptor.

The rest of the paper is organized as follows. The related literature will be reviewed in Section 2. In Section 3 the basic required elements to describe the proposed method are presented. Section 4 presents the stereo camera feature descriptor algorithm. Section 5 gives the experimental results followed by the conclusions given in Section 6.

\section{Related Work}

A complete overview on visual odometry can be found in \cite{Scaramuzza2011} and \cite{Scaramuzza20112}. Feature descriptors have been used in visual odometry for many years. But descriptors became more important in this application after \cite{Nister2006}. They suggested using descriptors to match features between the left and right camera images and by using them to track features over time. This technique has recently become more popular and consequently, computing feature descriptors has become one of the major steps in the success of a visual odometry algorithm.

One of the well-known and robust feature descriptors is \textit{SIFT}. It has been successfully used for more than one decade in many applications including visual odometry, scene reconstruction, object recognition, etc. The \textit{SIFT} descriptor vector contains 128 floating point numbers. The main issue with the \textit{SIFT} descriptor is its computation and matching times. In real-time applications such as visual odometry or Simultaneous Localization and Mapping (\textit{SLAM}), time limitation is a serious challenge. Moreover, in \textit{SLAM}, it is needed to store feature descriptor vectors in order to find loop closure which is the main step of the path optimization. Therefore, the size of the descriptor vector is very important. The \textit{SURF} descriptor, which is a fast version of \textit{SIFT}, constructs a 64-D descriptor vector and reduces the computation time compared to the \textit{SIFT} descriptor. However, it still has high computation and matching time. \cite{mikolajczyk2005performance} showed that using dimensionality reduction techniques such as Principal Component Analysis (PCA) or Linear Discriminant Embedding (LDE) can reduce descriptor size without any loss in recognition performance. Another way to reduce the descriptor vector size has been presented by \cite{Tuytelaars2007}. They took advantage of a quantization method to use only 4 bits to store floating numbers of the descriptor.

Many methods have been employed to solve the \textit{SIFT} problems to reach a fast computation and matching as well as a suitable discriminative power. \textit{BRIEF} descriptor is one of the successful alternatives to \textit{SIFT}. It has a good performance and a low computation cost. \textit{BRIEF} computes a binary descriptor where each bit is independent. Therefore, the matching process would be much easier than the SIFT-like descriptors. Indeed, the Hamming distance would be used and the matching time will decrease significantly. More specifically, \textit{BRIEF} defines a test on some pairs of image patches around the feature. The pixel intensity of each smoothed patch is computed and the test is performed on them. The result of the test which might be true or false, determines one bit of the descriptor. In other words, the \textit{BRIEF} descriptor is an n-dimensional bit string, where each bit is the result of a test between two patches. They consider n=128, 256 and 512 in the largest configuration, and the vector needs just 64 bytes. A significant problem with \textit{BRIEF} is that it is neither rotation- nor scale-invariant.

The \textit{ORB} descriptor, uses the keypoint direction to steer the \textit{BRIEF} descriptor. In addition, a learning method is developed to choose a good subset of binary tests. It is a rotation invariant version of the \textit{BRIEF} descriptor and it is highly robust to noise. \textit{FREAK} is another method which is quite similar to the \textit{ORB} descriptor. \textit{FREAK}, inspired by human visual system, creates a sampling pattern which is extracted from the human retina. In human eyes, the distribution of ganglion cells is not uniform. Their distribution decreases exponentially as the cells take distance from the foveal area. \textit{FREAK} uses the retina sampling pattern, which means that it raises the patch size according to the distance from the feature. In this way, descriptor extracts more detailed information from the area around the feature and extracts more global information as the patch takes distance from the feature. Moreover, the patches have some overlap in the receptive fields and lead to a better performance. \cite{verma2018local} combined local neighborhood difference pattern
(LNDP) and local binary pattern (LBP) to extract more information from local intensity of surrounding pixels of the feature. On the other hand, \cite{zhang2018feature} proposed a feature descriptor based on Normalized Difference Vector (NDV). It was able to takes full advantage of the local difference between the neighboring pixels. Moreover, they proposed two strategies to make it rotation invariant.

There are some approaches to construct the descriptor vector by learning a convolutional neural network such as \cite{Jahrer2008}, \cite{Osendorfer2013}, and \cite{Simo-Serra2015}. Most of these works tried to train a convolutional neural network works to extract the features of an input patch using an enormous amount of training patches. Then use these features as a descriptor. These descriptors are similar to the SIFT-like descriptors and The Euclidean distance reflects the features similarity. As a result, feature matching is more time-consuming than the binary descriptors.

\section{Preliminary}
In this section, the basic elements needed to describe the proposed method are presented. First, a brief description of features is provided. Second, the feature descriptor is explained, and finally, a basic visual odometry algorithm is briefly presented.

\subsection{Feature Point}
A feature point is a pixel or a small area in an image which differs from its immediate neighborhood in terms of intensity, color, or texture \cite{Scaramuzza20112} which in general, is called feature. The pixels around a good feature should have meaningful information in order to discriminate the feature from other points in the image. For example, a repeated plain texture or a simple edge is not a good feature. A good feature is a point in the 3D world which is projected on the camera image and it is easily locatable from different views. 

There are many feature extraction algorithms such as \textit{SIFT}, \textit{SURF}, \textit{FAST}, and \textit{Harris}. Each feature has some important properties including its \textit{location}, \textit{response} and \textit{size}. The \textit{location} shows the coordinate of the point in the image and \textit{response} is the score of the feature calculated by feature extraction algorithm. The other important property of a feature is its \textit{size} which represents the radius of the meaningful area around the feature.
\subsection{Feature Descriptor}
Feature descriptors have been used to match features between images of a scene taken from different views. Descriptors are numerical vectors constructed based on some calculation on the area around the feature. The goal of a descriptor is to extract meaningful information from the feature point to recognize it from different views. There are many feature descriptors for extracting this information and constructing the descriptor vector.

Most of the feature descriptor algorithms use a fixed circular area around the feature to construct the descriptor. Areas with different radii lead to different descriptors. Therefore, it is crucial to have reliable criteria to select the size of this area. Most of the feature descriptor algorithms use the \textit{size} properties of the feature to determine the area around the feature which should be used in the descriptor calculation.

Feature descriptors construct the descriptor vector with different characteristics. Some feature descriptors such as \textit{SIFT} and \textit{SURF} construct a descriptor vector containing floating point numbers. Another group of descriptors such as \textit{ORB} or \textit{FREAK} construct a binary descriptor. Binary descriptors have a very low time complexity in descriptor construction as well as distance computation between descriptors and matching.

\subsection{Visual Odometry}
The process of finding the translation between two positions of the camera based on images taken from those positions is called visual odometry. When this process is repeated on the images of a moving camera, the trajectory of the camera movement can be calculated. Many different visual odometry models are presented. All of the visual odometry models contain three main steps: feature extraction, feature matching or feature tracking, and motion estimation. In a stereo visual odometry model, first, the features of the left and right camera images are extracted. Then in the feature matching step, these features are matched for two time steps. Matching features between the left and right camera images and using triangulation methods results in 3D points. In the motion estimation step, these 3D points are used to estimate the motion of the camera between the two time steps.

\section{Stereo Feature Descriptor}
Some steps of the stereo visual odometry such as feature extraction and triangulation are mandatory. In the stereo visual odometry algorithms, 3D features are used to estimate the motion. We believe that these 3D features and the stereo images can be useful in constructing a more robust feature descriptor. 

We propose two strategies to use these data to improve the feature descriptor. First, we estimate the scale of each feature using the 3D information and feature distance from the camera. Second, we use the two images of the stereo cameras to construct the descriptor vector. These approaches can be applied to any existing feature descriptor.

\begin{figure}[!t]
\begin{center}
	
	\includegraphics[width=.8\linewidth]{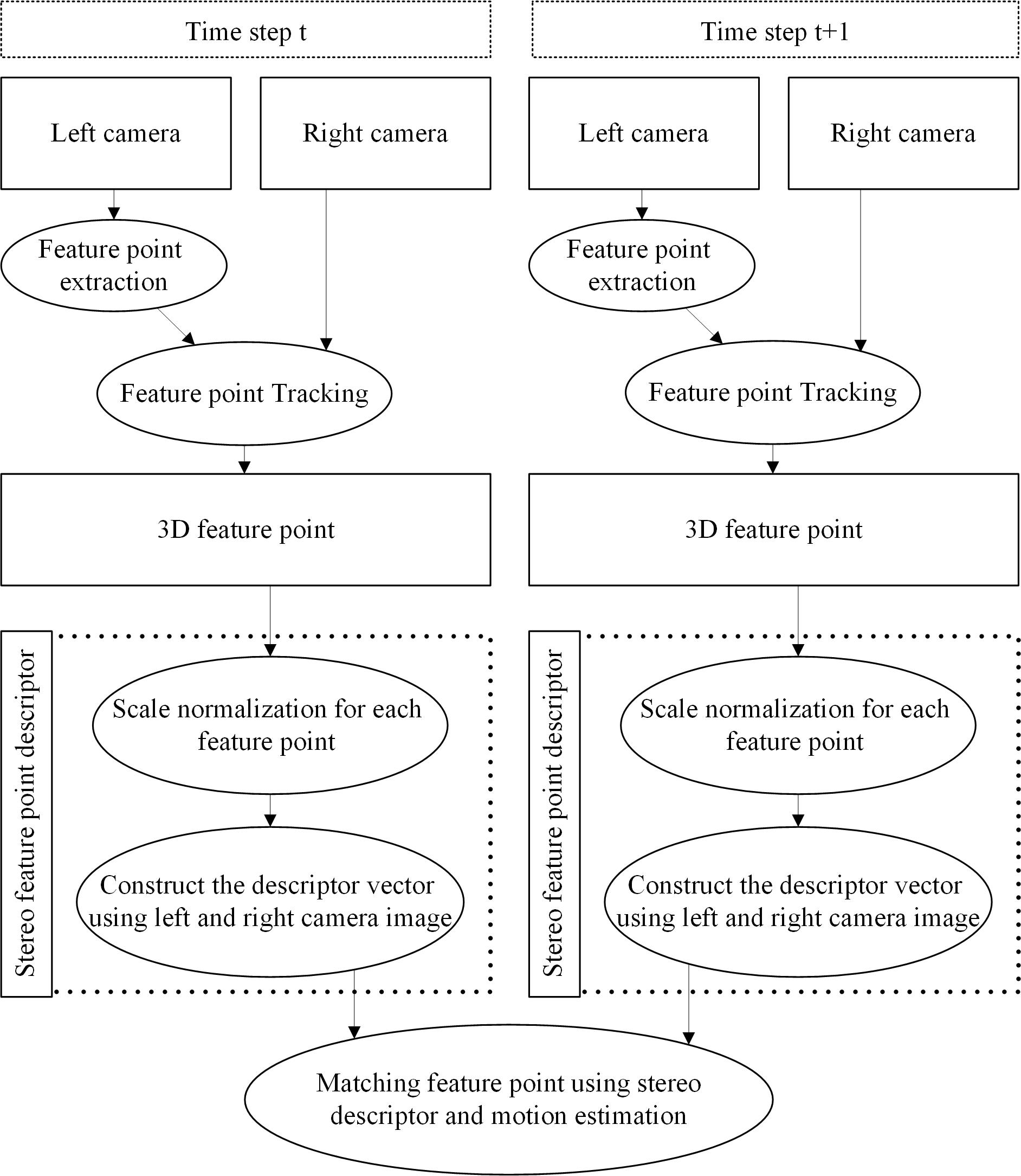}
\end{center}
\caption{Suggested visual odometry model to use the stereo feature descriptor.}
\label{fig:fig1}

\end{figure}

\begin{figure*}[t]
\begin{center}
	\includegraphics[width=.8\textwidth]{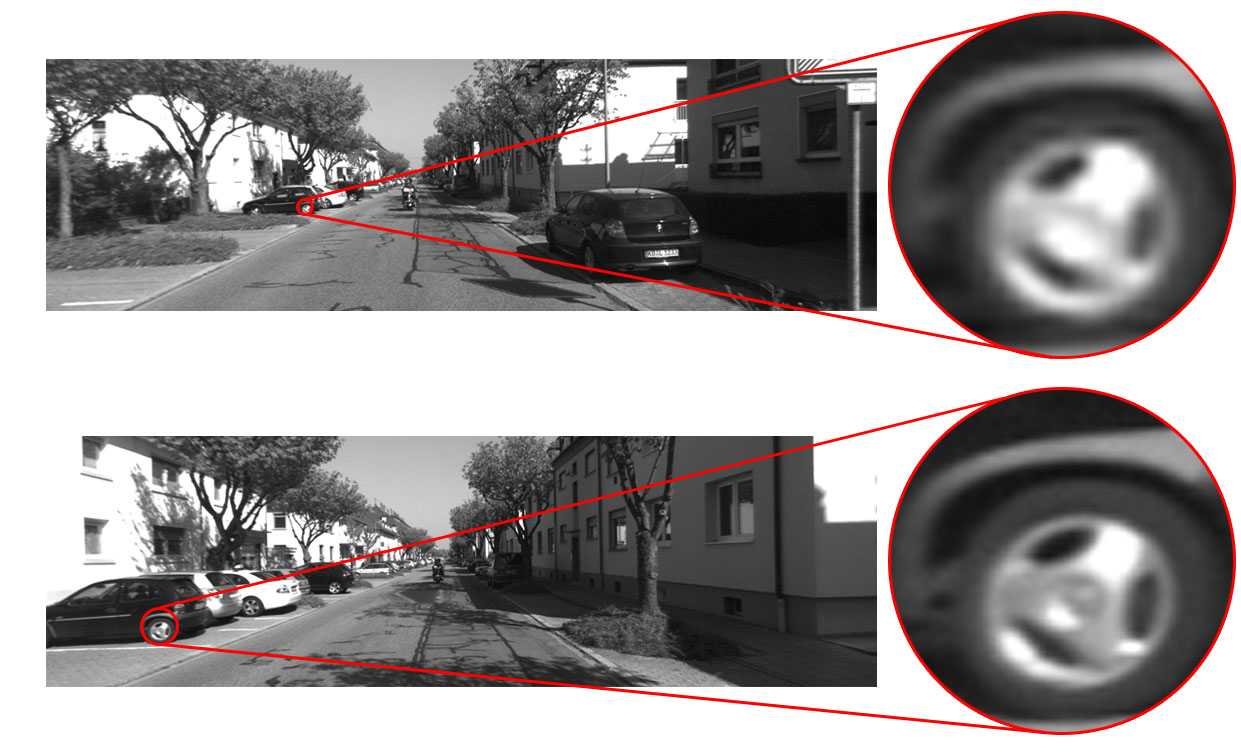}
\end{center}

\caption[width=\textwidth]{Finding the appropriate area around the feature based on its depth. The top picture shows a feature at a long distance from the camera and the bottom picture shows the same feature getting closer to the camera. By considering the depth information of the camera, the resulting areas show the same region.}
\label{fig:fig2}
\end{figure*}

The proposed stereo feature descriptor, requires 3D feature points to construct the feature descriptor. In the Figure \ref{fig:fig1}, a suggested visual odometry model is shown that calculate 3D features before the feature descriptors calculation. 
Each stereo visual odometry model has two feature matching step. First, between left and right camera images to construct 3D features. Second, between two consequent positions of the camera. 
In the suggested model, the first feature matching step is done using feature tracking technique which are accurate when there is a small changes in the camera positions. And the proposed stereo feature descriptor is used in the second feature matching step.

After feature matching, the depth information of the feature points is extracted. This is possible using several stereo geometry methods which is presented in  \cite{Hartley2003}. Since the depth information is available, the stereo feature descriptor can be constructed. Our stereo approach is divided into two steps. In the first step, the area around each feature is calculated based on its distance from the camera. The area is a circle where its radius is a function of the depth information of the feature point. The second step constructs the feature descriptor based on the left and the right camera images. The two steps are illustrated in the next subsections.

\subsection{Scale Normalization}

Pinhole cameras, like the human eye, follow the linear perspective projection rules. This means that objects are seen smaller as their distance from the camera increases. Therefore, the scale of an object in the image of the camera would change as the object changes its distance from the camera. Scale-invariant feature descriptors can solve this problem at the cost of intensive calculations. However, they are limited on the scale change in their algorithms. The time limitation in visual odometry forces the researchers to ignore the scale problem and use fast descriptors which are scale sensitive. In addition, we need to find loop closures to optimize the motions estimation. This process needs strong feature descriptors to recognize a previously visited location.

Nevertheless, there is some pre-calculated information which could help us to estimate the scale of features. The 3D positions of the features are used to estimate the motion of the camera. Therefore, a question arises: why not use this information to estimate the scale of a feature?

The calculation of the exact distance of the features from the camera is possible using the stereo images where a feature is viewed from two calibrated cameras. These calculations are called triangulation. In the proposed feature descriptor, we leverage the depth information of the features to determine a fixed area around the feature point in the image and construct the feature descriptor based on this area. In other words, even if a feature is seen from a different distance, the image portion which is used for the calculation of the descriptor would be the same. To do so, the feature descriptor is calculated on a circle around the feature where the radius of this circle is derived based on the depth of that feature. The relationship between the depth and the radius should be estimated. This can be done by experimentally sampling and applying a curve fitting algorithm.

Figure \ref{fig:fig2} demonstrates an example of such situation where a feature is viewed in two images with different distances from the camera. Here, a circle is drawn around the feature. This circle also shows the area used to construct the feature descriptor. Since we determine an image portion around the feature according to its depth, the same patterns would be used to construct the feature descriptor vectors and therefore, the descriptors would be the same in the two images.

\begin{figure}
\begin{center}
	\includegraphics[width=\linewidth]{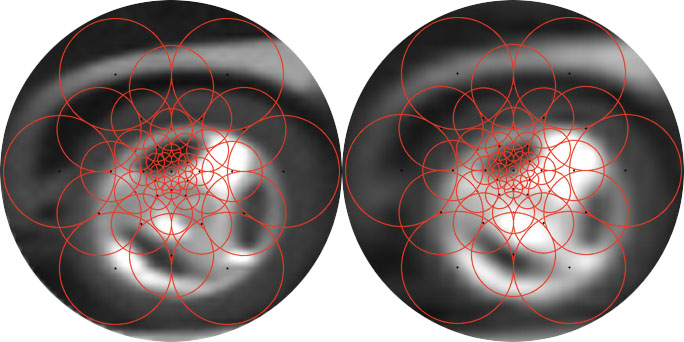}
\end{center}
\caption{Distribution of \textit{FREAK} Gaussian kernels around the feature in figure \ref{fig:fig2}. This shows that even if the two images have different resolutions, their kernel output would be the same. }
\label{fig:fig3}		
\end{figure}

The aforementioned policy can be applied to any descriptor by just modifying the \textit{size} property of the features. In this paper, we use \textit{FREAK} and \textit{SIFT} as the base descriptors to implement our stereo descriptors. The \textit{FREAK} descriptor uses Gaussian kernels, distributing them around the feature to extract local information. \textit{FREAK} is inspired from the distribution of visual cells in the human retina. Figure \ref{fig:fig3} shows these Gaussian kernels distributed in the two areas shown in Figure \ref{fig:fig2}. The two images are quite similar, but they are different just in terms of resolution and image quality. In this step, Gaussian kernels would help us, since changing resolution does not affect on its output.

\subsection{Stereo Descriptor}
Finding perfect matches between features is a crucial step of motion estimation in visual odometry. Among matched features, there are many outlier matches. Removing these outlier matches is possible by outlier rejection methods. After the rejection of outliers, motion estimation is applied to the inlier matches. However,  \cite{Cvi2015} proved that using all of the inliers would not lead to the best estimation. There are inliers which have some small errors. These errors are caused by small drifts in the feature extraction step. In addition, the feature descriptors are not sensitive enough to notice those small errors. As a result, they construct similar descriptors for these drifted points.

Our solution for this problem is to prevent matching drifted features by constructing a more strict feature descriptor. This goal is achieved using both images of the feature in the left and right cameras. These images have slight differences which are related to the angles of the two cameras. The feature is in the viewing field of both cameras, but the surrounding area of the point, in which the descriptor is made, is different.

Our descriptor vector has two sections. The first section represents the descriptor of the left view of the feature and the second section represents the descriptor of the right view of it. In this way, we have a double check on feature matching and prevent the feature matcher to match drifted features.
\section{Experimental Results}
The main aim of this paper is presenting a more robust feature descriptor for visual odometry, where 3D data is available. Therefore, the well-known visual odometry dataset called \textit{KITTI} presented by \cite{Geiger2012} is used to evaluate the proposed descriptor. Note that most datasets common for evaluation of feature descriptors, such as the dataset presented by \cite{mikolajczyk2005performance}, are monocular and do not support the stereo information. Thus they are not applicable to our stereo descriptor. Three experiments are designed to evaluate the proposed descriptor comprehensively. In the first experiment, the proposed scale normalized feature descriptor is evaluated. The second experiment examines the robustness of the stereo feature descriptor and the last experiment tests the stereo feature descriptor in a usual visual odometry problem. These experiments are presented in the next three subsections.

\begin{flushleft}
\textbf{ A. Feature Matching Experiment}
\end{flushleft}

The goal of this test is to show the degree of improvement the proposed scale normalization technique makes in the resulting descriptor. Therefore, this would be a contest between a standard descriptor and the scale normalized version of it. \textit{SIFT} and \textit{FREAK} descriptors are chosen for this evaluation. \textit{SIFT} is a scale invariant and robust descriptor and \textit{FREAK} is a fast and robust binary descriptor. We expect that the scale normalization technique would improves both of these descriptors.

In this experiment, we count the number of frames in which a feature could be found using the specified descriptor. It is obvious that a higher correctly matched score shows higher robustness of the descriptor. Here, each feature will be matched to features in the next ten following frames. In other words, this is a multi-step feature matching and the features of $frame\ t$ will be matched with the features of $frame\ (t + step)$, where:

\begin{center}
$ step\in\{1,2,3,...\ ,10\} $ and $ t=Current\ frame\ number $.
\end{center} 
This test is employed on the candidate descriptors two times. First, for the scale normalized version where the \textit{Scale Estimation} flag is set to \textit{True} and second, for the standard version where the \textit{Scale Estimation} flag is set to \textit{False}. Then, the evaluation algorithm operates as shown in Algorithm \ref{algorithm:al1}. 

\begin{algorithm}
\caption{Evaluation algorithm.}
\label{algorithm:al1}
\hspace*{\algorithmicindent} \textbf{Input:} The set of all points \\
\hspace*{\algorithmicindent} \textbf{Output:} The model that has the most consensus 
\begin{algorithmic}[1]
	\State {Extract features from $ frame=t $ and $ frame=t+step $.}
	\State {Calculate depth information of each feature using left and right camera images and calibration data.}
	\State {Estimate the scale of each feature and reset the \textit{size} properties for each feature.}
	\State {Calculate the descriptors using standard algorithm and match features.}
	\State {For each correct match the descriptor gets one point.}
	\State {if $ step<=10 $, $ step=step+1 $ and repeat from Step 1.}
	\State {$ t=t+10 $, $ step=1 $ and repeat from Step 1.}
	
\end{algorithmic}
\end{algorithm}

Table 1 shown that scale normalized versions of \textit{SIFT} and \textit{FREAK} descriptors, named \textit{SN-SIFT} and \textit{SN-FREAK}, both got better scores in all sequences than their standard versions. \textit{SIFT} is a scale invariant descriptor, so as it was expected, the improvement in \textit{FREAK} is higher than in \textit{SIFT}. However, the scale normalized \textit{SIFT} also had a 8.75\% improvement on the average. Since \textit{Freak} is not scale invariant, the proposed technique helps \textit{FREAK} to improve its score by 28.65\% on the average. Scale normalization helps descriptors to become scale invariant feature descriptors.

\begin{table}
\caption{The earned scores by each descriptor in each sequence of \textit{KITTI} dataset (* \textit{SN}: Scale normalized version).}
\begin{center}
	\resizebox{\columnwidth}{!}{%
		\begin{tabular}{ |c| c| c| c |c |c| c |}
			\hline

			& \multicolumn{3}{c}{\textit{FREAK}} 
			& \multicolumn{3}{|c|}{\textit{SIFT}} \\
			\cline{2-7}
			Descriptors: & \textit{SN}* & Standard & \% Improvement & \textit{SN}* &  Standard & \% Improvement  \\
			\hline\hline
			Sequence 0 & 1123 & 868 & 29.34\%  & 1398 & 1260 & 10.95\%  \\
			Sequence 1 & 2697 & 2278 & 18.42\%  & 1752 & 1627 & 7.65\%  \\
			Sequence 2 & 2671 & 2074 & 28.81\%  & 1847 & 1692 & 9.16\%  \\
			Sequence 3 & 2184 & 1784 & 22.39\%  & 1580 & 1483 & 6.53\%  \\
			Sequence 4 & 927 & 701 & 32.25\%  & 878 & 818 & 7.37\%  \\
			Sequence 5 & 2481 & 1842 & 34.69\%  & 1706 & 1556 & 9.61\%  \\
			Sequence 6 & 2143 & 1671 & 28.23\%  & 1663 & 1553 & 7.04\%  \\
			Sequence 7 & 2517 & 1870 & 34.64\%  & 1101 & 1024 & 7.46\%  \\
			Sequence 8 & 2627 & 1926 & 36.4\%  & 1899 & 1680 & 13.05\%  \\
			Sequence 9 & 1057 & 849 & 24.53\%  & 1150 & 1063 & 8.15\%  \\
			Sequence 10 & 1603 & 1279 & 25.34\%  & 1236 & 1151 & 7.34\%  \\
			Sequence 11 & 2022 & 1631 & 23.97\%  & 1402 & 1319 & 6.32\%  \\
			Sequence 12 & 1715 & 1254 & 36.68\%  & 1582 & 1425 & 11\%  \\
			Sequence 13 & 2338 & 1766 & 32.35\%  & 1902 & 1731 & 9.86\%  \\
			Sequence 14 & 2953 & 2460 & 20.06\%  & 1990 & 1857 & 7.17\%  \\
			Sequence 15 & 1040 & 799 & 30.1\%  & 1188 & 1084 & 9.59\%  \\
			Sequence 16 & 2612 & 2089 & 25.02\%  & 1925 & 1788 & 7.65\%  \\
			Sequence 17 & 2282 & 1712 & 33.28\%  & 1874 & 1691 & 10.8\%  \\
			Sequence 18 & 3839 & 3060 & 25.44\%  & 3100 & 2860 & 8.39\%  \\
			Sequence 19 & 2216 & 1629 & 35.99\%  & 1640 & 1496 & 9.58\%  \\
			Sequence 20 & 1259 & 999 & 25.94\%  & 987 & 903 & 9.38\%  \\
			Sequence 21 & 2263 & 1790 & 26.41\%  & 1906 & 1756 & 8.51\%  \\

			\hline
		\end{tabular}
	}
\end{center}
\label{table:tb1}

\end{table}

Feature extraction algorithms often use a threshold on the \textit{response} property of the features to prune weak features. Using a strict threshold results in a small set of features with a high \textit{response} property. In another word, a higher \textit{response} indicates a stronger feature. Therefore, with the use of a strict threshold, it would be easy for descriptors to find features in consecutive frames and the proposed method results in a limited improvement over standard descriptors. However, in this experiment, in order to show the advantage of the proposed method, we use a loose threshold, so that the feature set has both strong and weak features. The experiment was repeated three times. The feature set size for these experiments was 310, 1350, and 2320, respectively. Figure \ref{fig:fig7} shows the improvement of the proposed method over the standard descriptors for different sizes of the feature set. We see that the improvement increases as the size of the feature set increase.

\begin{figure}
\begin{center}
	\begin{tikzpicture}
	\begin{axis}[
	title={ },
	xlabel={Feature Set Size},
	ylabel={\% Improvement},
	ymin=0, ymax=30,
	legend pos=north west,
	ymajorgrids=true,
	grid style=dashed,
	]
	\addplot[color=blue,mark=square,]
	coordinates {(310,11.5819)(1350,15.37)(2320,26.4118)};
	\legend{SN \textit{FREAK} Improvement}
	\end{axis}
	\end{tikzpicture}
\end{center}
\caption{The improvement of the proposed method over standard \textit{FREAK} descriptor for different sizes of the feature set }
\label{fig:fig7}
\end{figure}
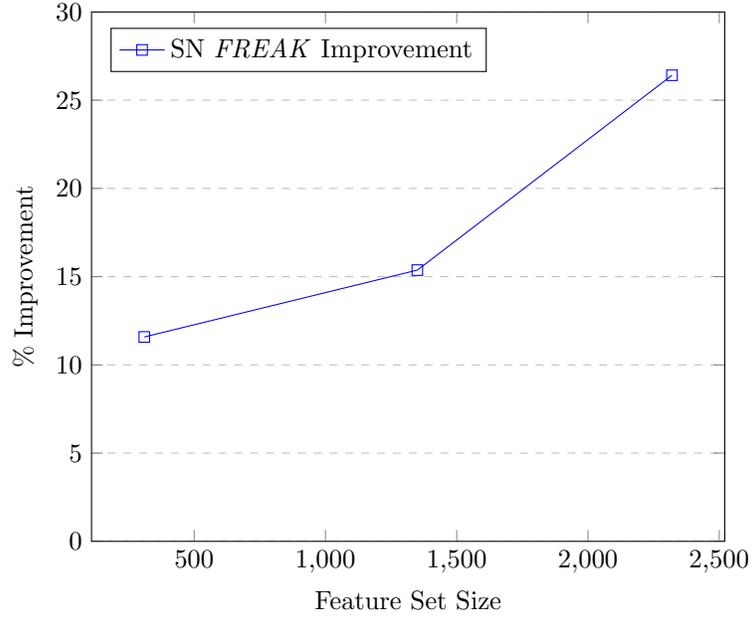
\begin{flushleft}
\textbf{ B. Visual Odometry Robustness Experiment}
\end{flushleft}
To evaluate the robustness of the proposed method, we design an inlier counting visual odometry experiment in which we use the simple visual odometry model presented in Figure \ref{fig:fig1}. In this test, instead of fitting the model to the dataset to get the highest accuracy, we aim to examine the robustness of the descriptors. Therefore, a multi-step visual odometry model is used where motion estimation will be applied to the current frame $ t $ and $ t+step $ frame later for $ step=\{1,2,3,4,5\} $. In a standard visual odometry, the $step$ is equal to one. When a larger $step$ is chosen, a larger transform would occur between frames and it becomes more challenging for the descriptor to cope with the scale change of each feature.
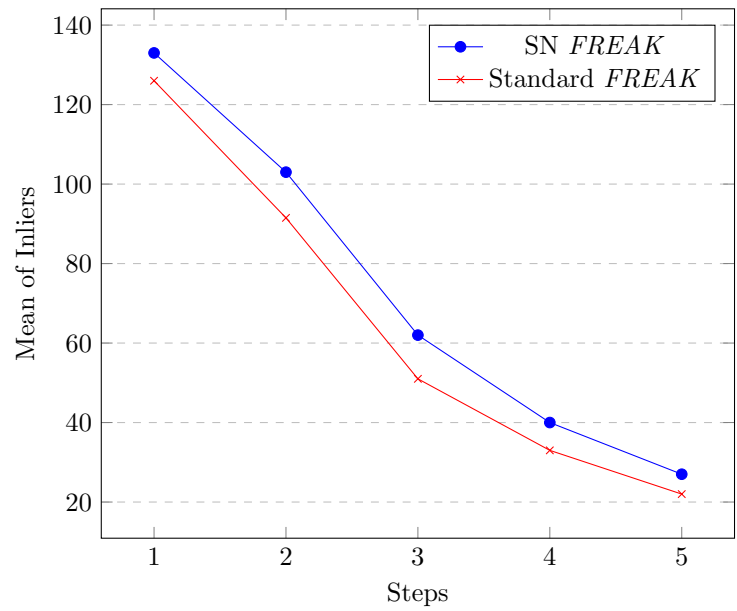
\begin{figure}
\begin{center}
	\begin{tikzpicture}
	\begin{axis}[
	title={ },
	xlabel={Steps},
	ylabel={Mean of Inliers},
	legend pos=north east,
	ymajorgrids=true,
	grid style=dashed,
	]
	\addplot[color=blue,mark=*,]
	coordinates {(1,133)(2,103)(3,62)(4,40)(5,27)};
	\addplot[color=red,mark=x,]
	coordinates {(1,126)(2,91.5)(3,51)(4,33)(5,22)};
	
	\legend{SN \textit{FREAK},Standard \textit{FREAK}}
	\end{axis}
	\end{tikzpicture}
\end{center}
\caption{The Average number of inliers in the visual odometry experiment on the \textit{KITTI} dataset. The x axis shows the steps and the y axis shows mean of inliers. }
\label{fig:fig5}
\end{figure}

\begin{table*}
\caption{The average result of 20 runs of visual odometry test on \textit{KITTI} dataset. Each cell represents the average translation error of all paths}
\begin{center}
	
	\begin{tabular}{>{\centering\arraybackslash}m{30mm}>{\centering\arraybackslash}m{25mm}>{\centering\arraybackslash}m{25mm}>{\centering\arraybackslash}m{25mm}@{}m{0pt}@{}}
		\hline
		Method & Best estimation & Average of estimation & Worse estimation& \\
		\hline\hline
		Stereo version of \textit{FREAK} & 2.47\% & 3.5\% & 4.29\%& \\
		\hline
		Standard \textit{FREAK} & 3.43\% & 4.78\% & 8.8\% & \\
		\hline
	\end{tabular}
\end{center}
\label{table:t2}

\end{table*}
We applied our stereo version of \textit{FREAK} descriptor on the $ KITTI $ dataset and compared it against the standard version of the \textit{FREAK} descriptor. The results of this test are shown in Figure \ref{fig:fig5}. Our stereo descriptor produces 19\% more inliers in comparison to the standard version of the \textit{FREAK} descriptor. Due to the strict criteria on matching features in our method, considering the number of inliers as the only measure cannot constitute a comprehensive and conclusive evaluation. In order to deduce a better evaluation, we also perform an accuracy based visual odometry test on the stereo feature descriptor algorithm.

\begin{flushleft}
\textbf{ C. Visual Odometry Accuracy Experiment}
\end{flushleft}
In the final experiment, the accuracy improvement is evaluated. As before, we used the simple visual odometry model illustrated in Figure \ref{fig:fig1}. For outlier rejection we used \cite{shojaedini2017novel}. This method has an important random subsection which causes the model to have different results in each run even with fixed parameters. One way to cope with the randomness is to repeat the test for several times with fixed parameters. Therefore, the test is repeated 20 times for each descriptor and the average results are reported to have a better comparison. As mentioned before, this is not a benchmarking test and we want to evaluate the general accuracy improvement of our method.

Repeating the experiment for several times, resulted in 23\% improvement in the average estimation of the standard FREAK by using our stereo \textit{FREAK} descriptor. Table \ref{table:t2} shows the result of all tests. The stereo version of \textit{FREAK} descriptor has better results in both worse and best estimations, too. This shows that the stereo descriptor is more stable than the standard version and the scale normalization method is successful to improve the visual odometry accuracy and stability at the same time.

\section{Conclusions}
Feature descriptors are very useful in computer vision. Many algorithms such as visual odometry or SLAM rely on these methods to match features. 
Stereo visual odometry requires some mandatory computations such as 3D features calculation. This information can be used to improve feature descriptors. Our stereo feature descriptor uses the 3D information of features to adjust the scale based on the features depth. In this way, the descriptor robustness is improved without a significant computation. Moreover, we use stereo images to construct the descriptor vector. It helps the descriptor to have comprehensive information about the features and construct better descriptors. Our implementation increases the number of inlier matches of the \textit{SIFT} and the \textit{FREAK} descriptors by 8.75\% and 28.65\% on the average. Moreover, this technique results in 23\% improvement in visual odometry due to preventing wrong matches and adjusting the scale of the feature descriptor .

\bibliography{ref}

\end{document}